\documentclass[journal]{IEEEtran}
\ifCLASSINFOpdf
   \usepackage[pdftex]{graphicx}
\else
\fi
\usepackage{amsmath}
\usepackage{amssymb}
\usepackage{enumerate} 
\usepackage{cite}
\usepackage{color}
\usepackage{soul}
\usepackage{tabularray}
\usepackage{stfloats}
\usepackage{gensymb}
\usepackage{multirow}
\usepackage{bm}
\usepackage{graphicx}
\usepackage{colortbl}
\usepackage{xcolor}
\usepackage{fontawesome5}
\usepackage{xspace}
\usepackage{booktabs}
\usepackage{mathtools}
\usepackage[colorlinks,linkcolor=blue]{hyperref}

\usepackage{algorithm}
\usepackage[noend]{algpseudocode}

\usepackage{multirow}
\usepackage{array}
\usepackage{soul}
\soulregister{\cite}7
\soulregister{\ref}7
\UseRawInputEncoding

\floatname{algorithm}{Algorithm}

\definecolor{Gray}{gray}{0.90}
\definecolor{mygray}{rgb}{.80,.80,.80}
\definecolor{myblue}{rgb}{0.94, 0.95, 1.0}
\definecolor{mylightblue}{rgb}{0.96, 0.97, 1.0}
\definecolor{forestgreen}{rgb}{0.0, 0.5, 0.0}
\definecolor{ashgrey}{rgb}{0.7, 0.75, 0.71}
\definecolor{darkorange}{rgb}{1.0, 0.55, 0.0}

 \def \airspatialbotlogo {\raisebox{-0.2\height}{\includegraphics[height=0.85\baselineskip]{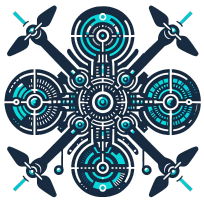}}}

\newcommand{\icoyes}{\textcolor{forestgreen}{\faIcon{check-circle}}\xspace}

\newcommand{\icono}{\textcolor{ashgrey}{\faIcon{times-circle}}\xspace}
\begin{document}

%

\title{\airspatialbotlogo{} AirSpatialBot: A Spatially-Aware Aerial \\ Agent for Fine-Grained Vehicle Attribute \\ Recognization and Retrieval}

%
%
%

\author{Yue~Zhou,
        Ran~Ding, 
        Xue~Yang,~\IEEEmembership{Member,~IEEE,}\\
        Xue~Jiang*,~\IEEEmembership{Senior Member,~IEEE,}
        and Xingzhao~Liu
\thanks{* Correspondence author.}
\thanks{Yue~Zhou, Ran~Ding, Xue~Jiang and Xingzhao~Liu are with the Department of Electronic Engineering, Shanghai Jiao Tong University, Shanghai 200240, China (e-mail: sjtu\_zy@sjtu.edu.cn;iyang0501@sjtu.edu.cn; xuejiang@sjtu.edu.cn;xzhliu@sjtu.edu.cn).}
\thanks{Xue~Yang is with the Department of Automation, Shanghai Jiao Tong University, Shanghai 200240, China (e-mail: yangxue-2019-sjtu@sjtu.edu.cn).}

}


%
%

\markboth{IEEE TRANSACTIONS ON GEOSCIENCE AND REMOTE SENSING}{Shell \MakeLowercase{\textit{et al.}}: Bare Demo of IEEEtran.cls for IEEE Journals}
%



\maketitle


\begin{abstract}

Despite notable advancements in remote sensing vision-language models (VLMs), existing models often struggle with spatial understanding, limiting their effectiveness in real-world applications. To push the boundaries of VLMs in remote sensing, we specifically address vehicle imagery captured by drones and introduce a spatially-aware dataset AirSpatial, which comprises over 206K instructions and introduces two novel tasks: Spatial Grounding and Spatial Question Answering. It is also the first remote sensing grounding dataset to provide 3DBB. To effectively leverage existing image understanding of VLMs to spatial domains, we adopt a two-stage training strategy comprising Image Understanding Pre-training and Spatial Understanding Fine-tuning. Utilizing this trained spatially-aware VLM, we develop an aerial agent, AirSpatialBot, which is capable of fine-grained vehicle attribute recognition and retrieval. By dynamically integrating task planning, image understanding, spatial understanding, and task execution capabilities, AirSpatialBot adapts to diverse query requirements. Experimental results validate the effectiveness of our approach, revealing the spatial limitations of existing VLMs while providing valuable insights. The model, code, and datasets will be released at \href{https://github.com/zytx121/AirSpatialBot}{https://github.com/VisionXLab/AirSpatialBot}.

\end{abstract}



%
\IEEEpeerreviewmaketitle

\section{Introduction}

\begin{figure*}[!t]
	\begin{center}             
        \includegraphics[width=1\linewidth]{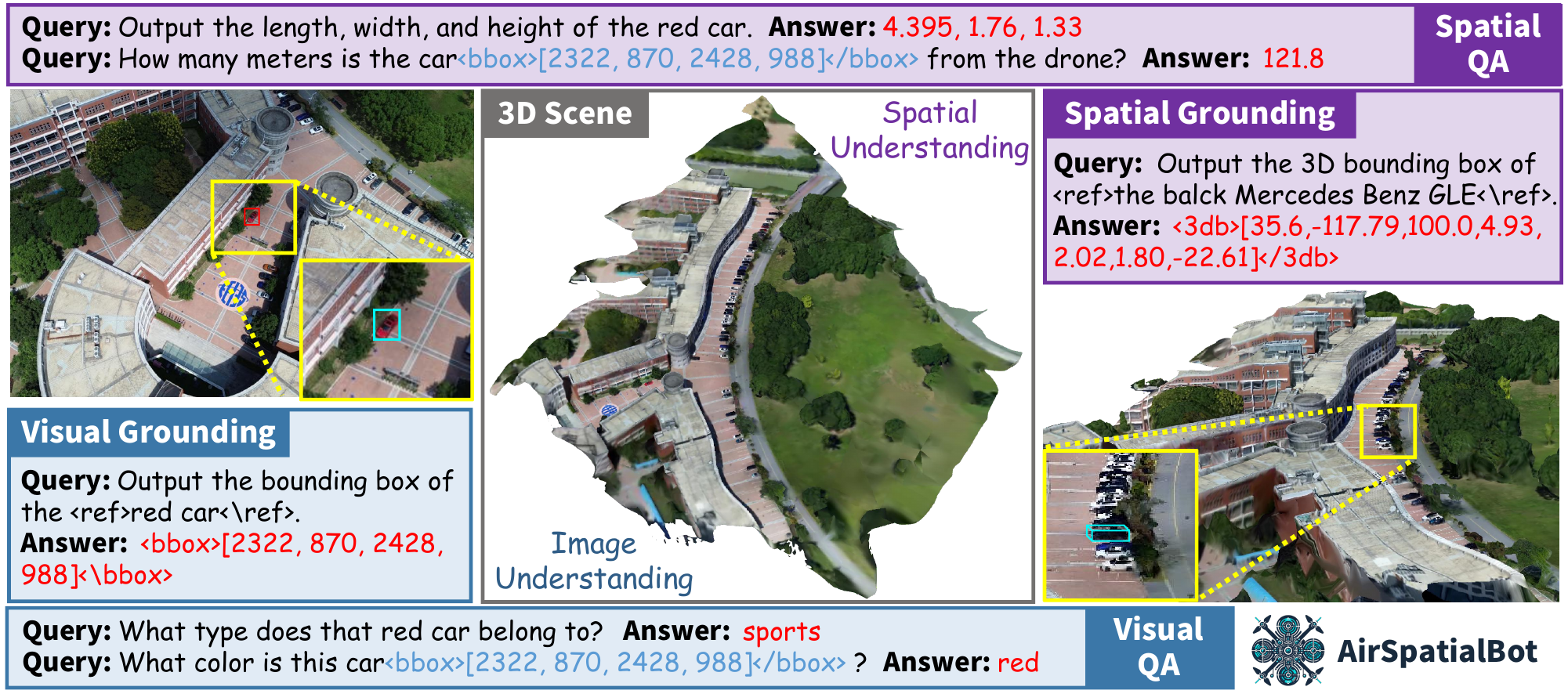}
    \end{center}
    \vspace{-0.5cm}
    \caption{Illustration of AirSpatialBot's Visual Understanding and Spatial Understanding Capabilities.}
    \label{fig:main}         
\end{figure*}

\IEEEPARstart{H}{umans} perceive the world by projecting 3D scenes onto the retinal plane, allowing them to interpret their surroundings and accurately estimate the distance between the object and their body, thereby avoiding collisions and potential injuries~\cite{yang2024thinking}. This ability, known as spatial understanding, facilitates the inference of 3D spatial information from 2D images and is fundamental to various applications, such as drone-based embodied intelligence~\cite{hu2024embod} and vision-language navigation~\cite{wang2024towards}, where aerial agents must accurately infer the 3D properties of objects from pictures captured by drones and take actions based on human instructions. However, despite demonstrating impressive visual understanding abilities, current Large Vision-Language Models (VLMs) in the remote sensing (RS) domain exhibit significant limitations in spatial relationship understanding~\cite{zhou2024towards}. 

To mitigate the spatial understanding limitations in VLMs, we introduce the AirSpatial dataset, which enhances 3D perception in aerial imagery. In addition to existing tasks such as visual grounding (VG)~\cite{qiao2021rec} and visual question answering (VQA)~\cite{Antol_2015_ICCV}, the dataset introduces two novel tasks: Spatial Grounding (SG) and Spatial Question Answering (SQA).  Fig. \ref{fig:main} illustrates the distinction between visual understanding and spatial understanding. Although both utilize 2D images, the following differences exist. VG relies on image-based referring expression (e.g., color, relative position), while SG employs spatially-aware descriptions (e.g., 3D dimensions, spatial distances). VG uses horizontal bounding boxes (HBBs) for evaluation, while SG can be evaluated using either HBBs or 3D bounding boxes (3DBBs). The SQA task consists of a series of regression problems, including depth, distance, length, width, and height, with numerical answers. It focuses on evaluating the model's understanding of spatial scales. The dataset contains over 80k question-3DBB pairs derived from 3D scenes reconstructed from multiple view aerial images~\cite{zhou2018open3d} captured by drones in low-altitude urban environments.

While AirSpatial provides a valuable dataset for spatial awareness, the availability of such data remains limited. To mitigate this constraint, we propose a two-stage training strategy that leverages existing 2D Remote Sensing Visual Grounding (RSVG) datasets~\cite{sun2022visual,zhan2023rsvg} to improve the model's comprehension of 3D spatial information. we first perform Image Understanding Pre-training on substantial 2D RSVG samples. Subsequently, we conduct Spatial Understanding Fine-tuning using AirSpatial data, equipping the model with explicit spatial grounding capabilities. Specifically, we employ auxiliary supervised learning (ASL) to facilitate knowledge transfer from 2D to 3D and introduce geometric mapping learning (GML) to ensure consistency by converting predicted 3DBBs in world coordinates back to HBBs in pixel coordinates. Through this process, we construct a remote sensing VLM with preliminary spatial awareness.

Based on the spatially-aware VLM, we develop a aerial agent, AirSpatialBot, that can perform fine-grained
vehicle recognition of brand, model, powertrain, and even price by estimating the vehicle's 3D dimensions (length, width, and height). In addition, it can retrieve vehicle with special attribute from the 3D scene. AirSpatialBot is composed of a Large Language Model (LLM) and our spatially-aware VLM. The LLM is responsible for generating a plan for each user query and invoking the VLM's capabilities to execute different plans, ultimately producing the answer. We envision that AirSpatialBot can unlock new possibilities for VLMs in remote sensing while providing researchers and practitioners with a novel approach
to aerial vehicle recognition. Unlike previous vehicle attribute recognition algorithms~\cite{Wang_2019_ICCV,li2021uav}, this approach does not require labeled images for every vehicle brand and model, significantly reducing annotation costs. When new vehicle models emerge, our approach allows easy adaptation to new vehicle models by simply updating the vehicle parameter table, eliminating the need for retraining.

\begin{table*}
\caption{Comparison Between Existing Remote Sensing Visual Grounding Datasets and Our AirSpatial-G.}
  \centering
  \setlength{\tabcolsep}{2pt}
  \begin{tabular}{lccccccccc}
    \toprule
    
    Dataset & Year & Publish & Source& \# Refers & GSD (m) & Image Width & HBB & OBB  & 3DBB \\

    \midrule
    RSVG~\cite{sun2022visual} & 2022 & ACM MM  & satelite & 5,505  & 0.24$\sim$4.88 & 1,024&   \icoyes & \icono & \icono   \\
    DIOR-RSVG~\cite{zhan2023rsvg} & 2023 & TGRS &  satelite  & 27,133  & 0.5$\sim$30 & 800 &  \icoyes & \icono & \icono  \\
    RSVG-HR~\cite{lan2024lqvg} & 2024 & TGRS &  satelite  & 2,650  & 0.24$\sim$4.88 & 1024 &  \icoyes & \icono & \icono  \\
    OPT-RSVG~\cite{li2024lgpa} & 2024 & TGRS &  satelite  & 48,952  & 0.15$\sim$30 & 152$\sim$10,569 &  \icoyes & \icono & \icono \\
   GeoChat~\cite{kuckreja2024geochat} & 2024 & CVPR & satelite  & 63,883  & 0.3$\sim$0.8 & 600$\sim$1,024 &  \icoyes & \icoyes  & \icono  \\
    VRSBench~\cite{li2024vrsbench} & 2024 & NeurIPS & satelite  & 38,689  & 0.1$\sim$30 & 512 &  \icoyes & \icoyes &\icono  \\    \rowcolor{myblue}AirSpatial-G 
    (Ours) & 2025 & - & drone & 80,497  & 0.007$\sim$0.04 & 4000 &   \icoyes & \icoyes  & \icoyes\\
    \bottomrule
  \end{tabular}
  \vspace{0.1cm}
    
    \label{tab:rsvg_dataset}
\end{table*}

In summary, our main contributions are as follows:

\begin{itemize}
  \item We introduce AirSpatial, a spatially-aware dataset featuring two novel tasks: Spatial
Grounding (SG) and Spatial Question Answering (SQA). It is the first RS grounding dataset to provide 3DBB, which will lead critical role of spatial understanding in RS VLMs.
  \item We propose a two-stage training strategy, pre-training on 2D RSVG datasets and fine-tuning with AirSpatial to enhance spatial understanding. To facilitate 2D-to-3D knowledge transfer, we introduce ASL, while GML ensures 3D spatial consistency.
  
  \item We develop AirSpatialBot, an aerial agent that utilizes our spatially-aware VLM for fine-grained vehicle attribute recognition and retrieval, making it the first approach capable of identifying vehicle brands, models, and pricing information from aerial imagery. 
  
  \item Extensive experiments demonstrate the effectiveness of AirSpatialBot, offering a new perspective on fine-grained target attribute recognition in aerial imagery.

\end{itemize}

\section{Related Work}

\subsection{Remote Sensing Visual Grounding}

RSVG aims to use natural language expression to locate specific objects in RS images, facilitating intuitive human interaction with intelligent RS interpretation systems. Compared to other multimodal RS tasks, such as image captioning~\cite{zhang2021global,li2021recurrent}, text-image retrieval~\cite{mikriukov2022deep}, and VQA~\cite{lobry2020rsvqa} in RS, RSVG is relatively novel and currently underexplored. As shown in Tab. \ref{tab:rsvg_dataset}, existing RSVG datasets typically provide 2D location signals, such as HBBs or oriented bounding boxes (OBBs). Moreover, their referring expressions are limited to the image plane (e.g., the relative position of objects such as top, bottom, left, or right), thus neglecting the spatial understanding capabilities of models. To address this limitation, this paper introduces the first RSVG dataset that provides 3DBBs as location signals, effectively extending object referring descriptions from 2D image planes to full 3D spatial contexts. For instance, targets can now be referenced based on their 3D dimensions or their distance from the camera.

\subsection{Remote Sensing Visual Question Answering}

Recently, multiple Remote Sensing Visual Question Answering (RSVQA) datasets ~\cite{lobry2020rsvqa,zheng2021mutual,zhang2023crsvqa} have been developed. While these datasets have advanced the application of VQA systems in drone-based tasks such as disaster assessment~\cite{maryam2021floodnet}, change detection~\cite{yuan2022cdvqa}, and urban planning~\cite{wang2024earthvqa}, they typically emphasize visual perception while neglecting explicit spatial reasoning. In addition, existing remote sensing datasets, despite containing a large number of ground targets, often suffer from coarse-level annotations, typically limited to fundamental attributes such as color or general type. This lack of detailed labeling severely restricts models' ability to recognize finer-grained attributes of objects, which constrains the variety and depth of remote sensing VQA tasks. As a result, the full potential of VLMs in remote sensing remains largely untapped. To address these limitations, we propose a spatially-aware VQA dataset, \textsc{AirSpatial-QA}. As shown in Tab.~\ref{tab:rsvqa_dataset}, it is the first RSVQA dataset explicitly incorporating three-dimensional spatial relationships of aerial targets, aiming to enhance the model's spatial understanding capabilities. Moreover, it covers fine-grained attributes of 11 types of vehicles, featuring a diverse range of questions that require specialized knowledge in the field of vehicles.

\begin{table*}
\caption{Comparison Between Existing Remote Sensing VQA Datasets and Our AirSpatial-QA.}
  \centering
  \setlength{\tabcolsep}{2pt}
  \begin{tabular}{lcccccccccc}
    \toprule
    Dataset  & Year & Publish & Source & \#Refers & Image Width & 2D-Pos. & 3D-Pos. & 3D-Size & Depth \\
    \midrule
    RSVQA-LR\cite{lobry2020rsvqa} & 2020 & TGRS  & satelite& 77,232  & 512   & \icoyes & \icono  \\
    RSVQA-HR\cite{lobry2020rsvqa} &2020 & TGRS& satelite&   1,066,316  & 512   & \icoyes & \icono & \icono & \icono \\
    RSVQAxBEN\cite{lobry2021rsvqaxben} & 2021 & IGARSS & satelite & 14,758,150  & 20$\sim$120  &\icono & \icono & \icono & \icono  \\
    FloodNet \cite{maryam2021floodnet} & 2021 & IEEE Access   & drone & 11,000 & 4,000   & \icono & \icono & \icono & \icono  \\
    RSIVQA \cite{zheng2021mutual} & 2021  &  TGRS &satelite & 111,134 &  256$\sim$4,000   & \icoyes & \icono & \icono & \icono  \\
    CDVQA \cite{yuan2022cdvqa} & 2022   & TGRS & satelite& 122,000  &  512   & \icono & \icono & \icono & \icono  \\
    VQA-TextRS \cite{al2022open} & 2022 & IJRS &satelite & 6,245  &  256$\sim$600   & \icono & \icono & \icono & \icono  \\
    CRSVQA \cite{zhang2023crsvqa}    & 2022 & TGRS & satelite& 4,644  & 600    & \icoyes & \icono & \icono & \icono  \\
    RSEval \cite{hu2023rsgpt}  & 2024 & ArXiv & satelite & 936  & 512    & \icoyes & \icono & \icono & \icono  \\
    
    EarthVQA \cite{wang2024earthvqa}   & 2024 & AAAI & satelite& 208,593  & 1,024   & \icoyes & \icono & \icono & \icono  \\
    VRSBench \cite{li2024vrsbench} &  2024 & NeurIPS & satelite &  123,221 & 512   & \icoyes & \icono & \icono & \icono  \\
    \rowcolor{myblue}\textsc{AirSpatial-QA (Ours)}  & 2025 & - & drone & 126,006  & 4,000   & \icono & \icoyes  & \icoyes & \icoyes  \\
    \bottomrule
  \end{tabular}
  \vspace{0.1cm}
    
    \label{tab:rsvqa_dataset}
\end{table*}

\subsection{Spatially-aware VLMs}

Current research on spatially-aware Vision-Language Models (VLMs) primarily focuses on embodied intelligence~\cite{cai2024spatialbot} and autonomous driving~\cite{wang2024omnidrive}, where the primary objective is collision avoidance. However, these approaches often overlook unique challenges inherent in remote sensing scenarios, such as the need for fine-grained object recognition and retrieval in complex aerial environments. In fact, this capability is equally critical for various low-altitude aerial vehicles, which must maintain robust spatial awareness of their surrounding 3D environment and prevent crashes. Current remote sensing VLMs have shown promising results in image understanding tasks such as scene classification, image captioning, VQA, and VG~\cite{kuckreja2024geochat,muhtar2024lhrs,pang2024h2rsvlm,zhang2024earthgpt}. However, the spatial understanding of remote sensing VLMs remains relatively underexplored. To address this research gap, our paper introduces the first spatially-aware remote sensing VLM capable of effectively supporting spatial grounding and spatial VQA tasks, further exploring practical application scenarios for such capability.

\section{AirSpatial Dataset}

\subsection{Data Collection}

Existing RS datasets~\cite{xia2018dota,li2020object} often lack sensor metadata and provide limited object attributes. Consequently, these datasets can only support the formulation of simple visual questions, such as locating a red car in the upper-left corner of an image. We employ drone-based data collection to construct a spatially-aware aerial dataset from scratch. This approach ensures comprehensive access to sensor parameters, allowing inversion of the three-dimensional coordinates of image targets based on imaging geometry principles. Data collection is divided into two parts: aerial imagery and ground video.

\begin{figure}[!t]
	\begin{center}             
        \includegraphics[width=1\linewidth]{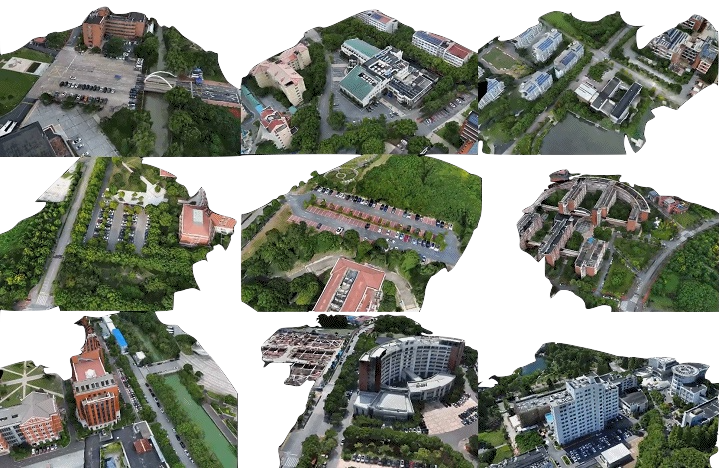}
    \end{center}
    \caption{The 3D scenes reconstructed using multi-view aerial images.}
    \label{fig:3d_scenes}         
\end{figure}

\paragraph{Aerial Imagery} 
All aerial images in AVVG are collected with a small UAV platform, DJI Mini3, in Shanghai. The dataset consists of 4K high-resolution RS images from 11 scenes, captured at 9 different above-ground levels and 3 pitch angles. As a result, these RS images exhibit varying spatial resolutions and perspectives. These scenes encompass diverse weather conditions and lighting scenarios. As shown in Fig. \ref{fig:3d_scenes}, we reconstructed these 3D aerial scenes using the multiview stereo reconstruction tool \cite{seitz2006mvs}.The first row consists of three scenes depicting cloudy weather, while the six scenes below showcase sunny conditions.

\paragraph{Ground Video} We record ground videos from the same areas to facilitate accurate annotation of vehicle brands and models. Specifically, we select time slots with relatively low vehicular mobility, avoiding rush hours and meal times. Additionally, to mitigate the vehicle mismatch between drone images and ground videos caused by vehicle entry and exit, we capture two sets of ground videos before and after the drone captures aerial photos. This ensures that vehicles entering or exiting the scene halfway through the capture are recorded in the videos. However, there are instances where vehicles pass through the scene briefly, leading to cases where they are not captured in either video. In such situations, we mask these vehicles with a black mask in the images to ensure that all visible vehicles have fully known attributes. Due to privacy concerns, ground videos will not be released.

\subsection{Metadata Annotation}
The metadata can be categorized into two main components. The first includes camera-related parameters, such as intrinsic parameters (focal length, pixel size, sensor dimensions) and extrinsic parameters (pitch angle, AGL). These are extracted from the raw data from the drone. The second component pertains to vehicle fine-grained attributes, which require manual annotation. To accurately describe the length and width of vehicles, we use rotated bounding boxes to annotate their positions~\cite{yang2022detecting}. Identifying specific vehicle brands from aerial imagery presents a significant challenge for human annotators, and as a result, existing publicly available RS vehicle datasets have not achieved brand-level annotations~\cite{mundhenk2016large,zhu2021detection}.

\begin{figure}[!t]
	\begin{center}             
        \includegraphics[width=1\linewidth]{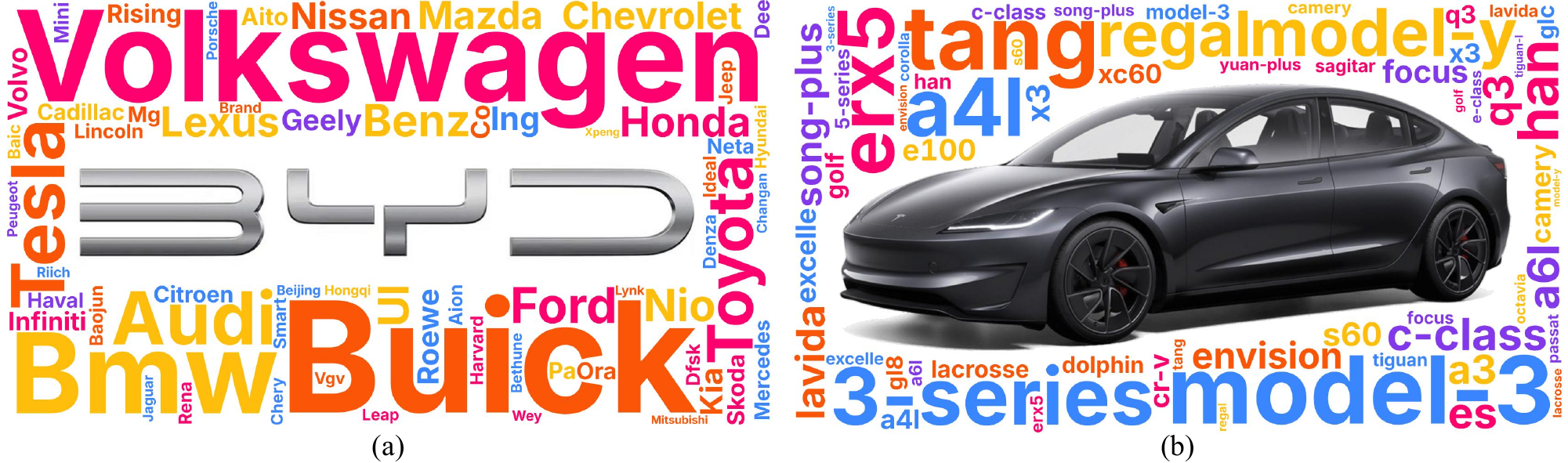}
    \end{center}
    \vspace{-0.5cm}
    \caption{Word cloud visualizations of vehicle occurrence frequencies in our dataset. (a) shows the brand word cloud, where BYD ranks first. (b) illustrates the model word cloud, with Tesla Model 3 ranking first.}
    \label{fig:clould}         
\end{figure}

\paragraph{Accessing to Fine-grained Vehicle Attributes}

Leveraging the previously mentioned ground videos, we successfully constructed the first RS vehicle dataset annotated with fine-grained attributes, identifying vehicles down to the model level within each brand. Specifically, we match the vehicles in the aerial image with the vehicles in the ground video one by one according to their locations and then call the DCD's API \footnote{\url{https://dcdapp.com}} to identify the specific model based on the vehicle's appearance and logo in the ground image. For vehicles whose models could not be identified, we used a black mask to cover them from the image. Then we used the DCD car database to obtain detailed attributes, such as the size and price of each car. Finally, we collected 814 vehicle instances, covering 53 brands and 211 models. Fig. \ref{fig:clould} presents a word cloud visualization of the occurrence frequencies of various brands and models. The top three most frequently occurring brands are BYD, Volkswagen, and Buick, while the top three most frequently occurring models are Tesla Model 3, BYD Song Plus, and Roewe ERX5. Moreover, our analysis shows that 32.3\% of the vehicles are new energy vehicles (NEVs), indicating widespread adoption of NEVs in China. These examples also underscore AirSpatialBot's potential for automotive market research applications.

\paragraph{Derivation of 3D Bounding Box}

\begin{figure}[!t]
	\begin{center}             
\includegraphics[width=1\linewidth]{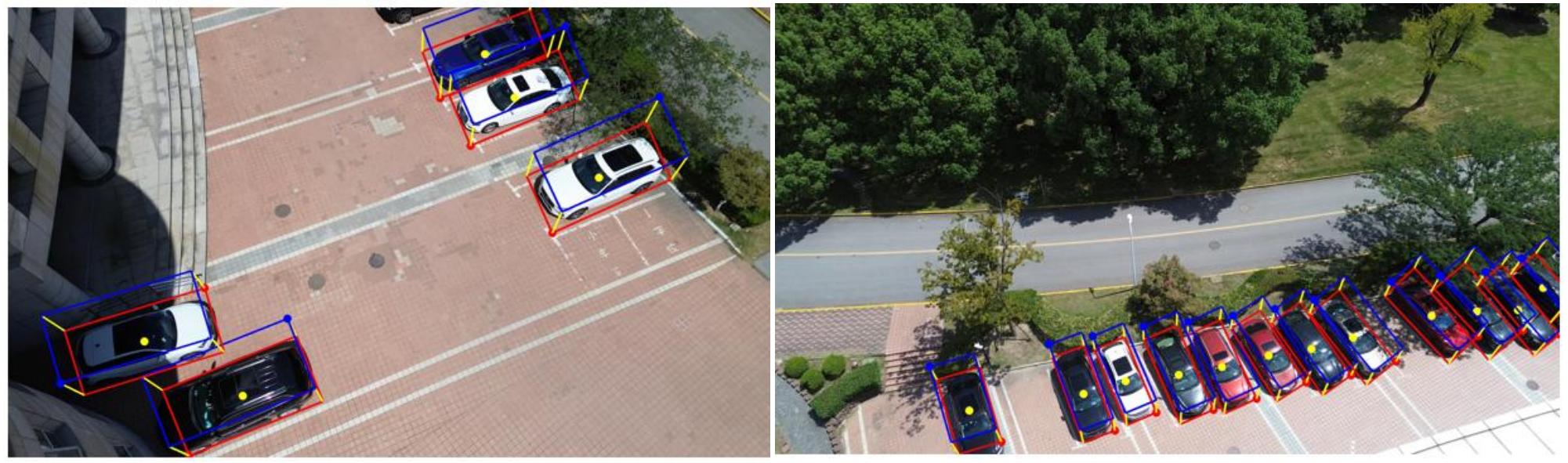}
    \end{center}
    \vspace{-0.5cm}
    \caption{Illustration of the 3D bounding box are obtained by Coordinate System Transformation.}
    \label{fig:3d_box}          
    \vspace{-0.3cm}
\end{figure}

Based on the previously annotated oriented bounding box of the vehicle, the 3D bounding box can be obtained through coordinate transformation. The transformation between the pixel coordinate system and the image coordinate system can be represented by an affine matrix, as follows:
\begin{equation}
\begin{split}
\left[ \begin{array}{c}
	x_P\\
	y_P\\
	1\\
\end{array} \right] =\left[ \begin{matrix}
	\frac{1}{p}&		0&		\frac{w}{2}\\
	0&		\frac{1}{p}&		\frac{h}{2}\\
	0&		0&		1\\
\end{matrix} \right] \left[ \begin{array}{c}
	x_I\\
	y_I\\
	1\\
\end{array} \right] 
\end{split}
\label{Equ:1}
\end{equation}
where $p$ represents the pixel size of the sensor. $\frac{w}{2}$ and $\frac{h}{2}$ denote the origin offsets, with the origin of the image coordinate system typically located at the image's top-left corner. Given the pixel coordinates of a certain point, its corresponding image coordinates can be calculated as follows:
\begin{equation}
\begin{split}
\begin{cases}
x_I = (x_P - w/2) \cdot p \\
y_I = (y_P - h/2) \cdot p
\end{cases}
\end{split}
\label{Equ:1}
\end{equation}

The transformation from the camera coordinate system to the image coordinate system is a conversion from three-dimensional to two-dimensional coordinates. Assuming the focal length of the camera is $f$, then we have
\begin{equation}
\begin{split}
z_c\left[ \begin{array}{c}
	x_I\\
	y_I\\
	1\\
\end{array} \right] =\left[ \begin{matrix}
	f&		0&		0&		0\\
	0&		f&		0&		0\\
	0&		0&		1&		0\\
\end{matrix} \right] \left[ \begin{array}{c}
	x_C\\
	y_C\\
	z_C\\
	1\\
\end{array} \right] 
\end{split}
\label{Equ:2}
\end{equation}
where $z_C$ denotes the depth of the point, which can be obtained by a depth camera (binocular or structured light). Because the drone camera we are using cannot provide depth information, we need to find another way.

When the ground satisfies the ground plane assumption, given the AGL of the drone and the pitch angle of the camera, the ground plane equation in the camera coordinate system is as follows:
\begin{equation}
\begin{split}
-\cos{\theta} \cdot Y_C - \sin{\theta} \cdot Z_C + H = 0 
\end{split}
\label{Equ:3}
\end{equation}

The equation of the line connecting the camera origin to the projection point on the pixel plane in the camera coordinate system is given by:
\begin{equation}
\begin{split}
\begin{cases}
X_C = x_I \cdot t \\
Y_C = y_I \cdot t\\
Z_C = f \cdot t
\end{cases}
\end{split}
\label{Equ:4}
\end{equation}

Substituting the line equation into the ground plane equation yields:
\begin{equation}
\begin{split}
t = \frac{H}{y_I\cos{\theta} + f\sin{\theta} }
\end{split}
\label{Equ:5}
\end{equation}

Substituting $t$ back into the line equation yields:
\begin{equation}
(\frac{x_I  H}{y_I\cos{\theta} + f\sin{\theta}}, \frac{y_I  H}{y_I\cos{\theta} + f\sin{\theta}}, \frac{f  H}{y_I\cos{\theta} + f\sin{\theta}})
\label{Equ:5}
\end{equation}

Fig. \ref{fig:3d_box} visualizes the 3D bounding box of a vehicle. The implementation procedure unfolds as follows: First, we calculate the 3D coordinates of the vehicle's center point within the camera coordinate system. Subsequently, based on the orientation information and the known vehicle dimensions (length, width, and height), we compute the 3D coordinates of its eight corner points. These computed 3D points are then projected back onto the 2D pixel plane. Finally, we manually refine the 3D bounding box to ensure it fully encompasses the target, which causes its dimensions to exceed the actual vehicle size. The visualization results further validate the correctness of this coordinate transformation. With accurate 3D bounding boxes established for each vehicle, we can effectively formulate challenging visual grounding tasks that explicitly demand spatial understanding capabilities from the model. Examples of such spatial tasks include identifying which vehicle is farthest from the camera or determining which vehicle possesses the greatest height.

To preserve the spatial mapping between camera coordinates and pixel coordinates, we refrained from cropping the 4K images to increase the dataset size, as is commonly done in most remote sensing datasets.

\subsection{AirSpatial-G Construction}

To extend the grounding capabilities of remote sensing VLMs from 2D to 3D, we construct a remote sensing spatial grounding dataset. This dataset is a specialized form of visual grounding, where all referential descriptions are inherently spatially oriented. Table \ref{tab:rsvg_dataset} summarizes the statistics of the existing remote sensing VG datasets. AirSpatial-G comprises 80k image-text-location pairs, with 66k in the training set and 14k in the test set. Following LLaVA\cite{liu2023llava}, we develop an instruction set based on AirSpatial-G to fine-tune VLMs. For each referred object, we provide three bounding box formats: HBB, OBB, and 3DBB, with five query templates for each format.

\subsection{AirSpatial-QA Construction}

AirSpatial-QA is an additional instruction dataset designed to enhance the spatial understanding of VLMs in aerial images. As shown in the blue dialog boxes in Fig. \ref{fig:main}, each sample in AirSpatial-QA comprises an aerial image and a single-turn dialog, containing spatial perception-related questions regarding the object's 3D dimensions, spatial distances, and depth information. Based on metadata, AirSpatial-QA is designed with five tasks, including estimating the depth, distance, and length-width-height information of the specified target. All tasks are formulated as open-ended questions, with correctness determined by constraining the error between the estimated value and the ground truth (GT) within 5\%. AirSpatial-QA generated a total of 126k image-question-answer pairs using templates, with 108k pairs in the training set and 17k pairs in the test set.

\subsection{AirSpatial-Bench Construction}

AirSpatial-Bench is a benchmark specifically designed for vehicle attribute recognition and retrieval tasks. It requires models to integrate planning ability, image understanding, and spatial understanding. Despite its high difficulty level, it offers substantial practical application value. We design two tasks for two application scenarios: (1) Vehicle Attribute Recognition: Users provide the 2D location of a vehicle in aerial imagery to query specific attributes of the vehicle, such as color, type, brand, model, drivetrain, number of seats, number of doors, and price. (2) Vehicle Retrieval: Users search for specific vehicles in aerial imagery based on provided brand and model information. The former task is essentially a visual question-answering problem, but recognizing attributes such as vehicle brand and model requires the model to leverage spatial understanding. The latter task is essentially a visual grounding problem, but we expect the model not only to locate the vehicle in the image but also to output its position in the 3D scene, enabling users to find their vehicle more intuitively. To ensure the differentiation of experimental results, we selected aerial images captured at relatively low altitudes (20 to 30 meters) to reduce task difficulty. Ultimately, we meticulously create 934 and 839 questions for two tasks, respectively.

\section{Spatial-Aware VLM}

Our spatial-aware VLM, inspired by LLaVA\cite{liu2023llava}, is composed of three core modules: a vision encoder, a projection layer, and a LLM. The vision encoder transforms aerial images into concise visual representations. The LLM integrates both textual and visual inputs to carry out reasoning tasks. To address the inherent modality gap between image data and textual understanding in LLMs, the projection layer acts as a crucial intermediary. To address the limitation of 3D data, we propose a transfer learning strategy that divides the training process into two phases: image understanding pre-training and spatial understanding fine-tuning. In the first phase, 2D visual grounding data and object detection data are used to align the model's understanding of details in remote sensing images. In the second phase, the 3D data proposed in this work is employed to further enhance the model's spatial understanding capabilities. The following subsections will introduce these two phases separately.

\subsection{Image Understanding Pre-training}

To achieve fine-grained alignment between remote sensing images and text, we collected four existing remote sensing visual grounding datasets~\cite{sun2022visual,zhan2023rsvg,kuckreja2024geochat,li2024vrsbench} and converted them into a unified format. In the end, a total of 187k referring descriptions with corresponding horizontal or rotated bounding box annotations were obtained. Each referring expression uniquely corresponds to one target in the image. Moreover, we have also incorporated extra instruction sets constructed from object detection samples sourced from the DOTA~\cite{xia2018dota}, DIOR~\cite{li2020object}, and FAIR1M~\cite{sun2022fair1m}. Specifically, we first cropped the original images from the FAIR1M and DOTA datasets into 512$\times$512 patches, while retaining the original image sizes for the DIOR dataset. We then used the object detection annotations from these datasets to generate training samples, where queries were formulated based on the categories present in each image. If an image contains objects from three categories, we constructed three image-text sample pairs accordingly. These data are used to enhance AirSpatialBot's 2D visual understanding capabilities.

Given an image $I$ and a referential text $x$, the model is required to integrate information from both the image and the text to perform image understanding, ultimately locate the coordinates $p^{2d}$ (e.g., HBBs or OBBs) of the referred object on the image plane. Thus, the training sample pairs in the first phase can be represented as:
\begin{equation}
    \mathcal{D}_{2d} = \{(I_i, x_i, p^{2d}_i)\}_{i=1}^{|\mathcal{D}|}
\end{equation}
We then train the model $\mathcal{M}$ on the $\mathcal{D}_{2d}$ using supervised fine-tuning (SFT) with a negative log-likelihood objective:
\begin{equation}
    \mathcal{L}_{phase1} = \; - \sum_{\mathcal{D}_{2d}} \; \log \mathcal{M}(y^{2d} \mid x, I),
\end{equation}
where the $y^{2d}$ is the  2D postion predicted by the model.

\begin{figure}[!t]
	\begin{center}             
\includegraphics[width=1\linewidth]{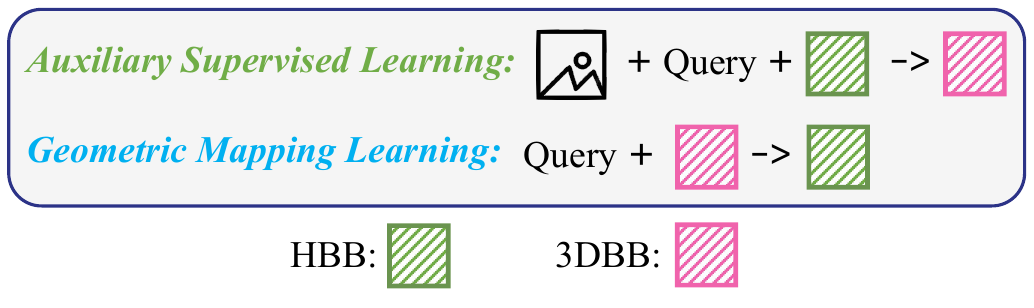}
    \end{center}
    \vspace{-0.5cm}
    \caption{Illustration of ASL and GML in the Spatial Understanding Fine-tuning.}
    \label{fig:asl_gml}          
    \vspace{-0.3cm}
\end{figure}

\subsection{Spatial Understanding Fine-tuning}

In the second phase, the model is required to integrate information from both the image and the query to perform spatial understanding, ultimately deriving the 3D coordinates $p^{3d}$ of the referential object in the 3D scene. However, due to the difficulty in collecting high-quality 3D visual localization data based on aerial images, constructing a large-scale set of $(I, x, p^{3d})$ pairs poses a significant challenge. This limitation hinders the enhancement of VLMs' spatial understanding capabilities through fine-tuning. To overcome this limitation, we utilize a small amount of data annotated with both 2D and 3D visual localization information $(I, x, p^{2d}, p^{3d})$ to efficiently transfer the VLM's visual localization capability from the 2D plane to 3D space through mixed-supervised fine-tuning. Specifically, we design two loss functions as illustrated in Fig. \ref{fig:asl_gml}, to assist the VLM in better understanding 3D based on 2D knowledge. The training set in the second phase can be represented as:
\begin{equation}
    \mathcal{D}_{3d} = \{(I_i, x_i, y^{2d}_i, y^{3d}_i)\}_{i=1}^{|\mathcal{D}|}
\end{equation}
Next, for each referred object, we train the model using instructions constructed based on both the 2D and 3D positions:
\begin{equation}
    \mathcal{L}_{Mix} = \; - \sum_{\mathcal{D}_{3d}} \; (\log \mathcal{M}(y^{2d} \mid x, I) + \sum_{\mathcal{D}_{3d}} \; \log \mathcal{M}(y^{3d} \mid x, I)
\end{equation}
We also propose a weakly supervised approach, where the 2D position of the referred object is provided as auxiliary information to encourage the model to predict the 3D position based on the 2D position. The ASL loss function is as follows:
\begin{equation}
    \mathcal{L}_{ASL} = \; - \sum_{\mathcal{D}_{3d}} \; \log \mathcal{M}(y^{3d}|x, I, y^{2d})
\end{equation}
Next, we propose enabling the model to learn the geometric consistency between 3D coordinates and their corresponding 2D coordinates, allowing the model to map 3D coordinates back to 2D coordinates without relying on the image. The GML loss function for this part is as follows:
\begin{equation}
    \mathcal{L}_{GML} = \; - \sum_{\mathcal{D}_{3d}} \; \log \mathcal{M}(y^{2d}|x, y^{3d})
\end{equation}
Finally, we obtain the overall loss for the second phase:
\begin{equation}
    \mathcal{L}_{phase2} = \mathcal{L}_{Mix} +
    \mathcal{L}_{ASL} + \mathcal{L}_{GML}
\end{equation}
 This training strategy maximizes the model's spatial understanding capability under conditions of limited 3D data availability, effectively mitigating issues caused by the scarcity of aerial 3D data.

\begin{table}
\caption{Overview of the Tools Utilized.}
\centering
\small
\renewcommand\tabcolsep{1.0pt} 
\begin{tabular}{cp{5.6cm}}
    \toprule
    \textbf{Type} & \textbf{Details} \\
    \midrule
    \multirow{2}{*}{\parbox{3cm}{\centering Spatial \\ \centering Understanding}}
    & Leverages the VLM to extract the 3D dimensions or spatial positions of vehicles. \\
    \midrule
    \multirow{2}{*}{\parbox{3cm}{\centering Image \\ \centering Understanding}}
    & Applies the VLM to perform visual analysis on image. \\
    \midrule
    \multirow{3}{*}{\parbox{3cm}{\centering Query Table}}
    & Retrieves additional attributes of vehicles from a database or infers thedimensions of a specific type of vehicle based on user input. \\
    \midrule
    \multirow{2}{*}{\parbox{3cm}{\centering Web Search}}
    & Supplements information not found in the database by performing a web search. \\
    \bottomrule
\end{tabular}

\label{tab:tools}
\end{table}

\begin{figure*}[!t]
	\begin{center}            
        \includegraphics[width=1\linewidth]{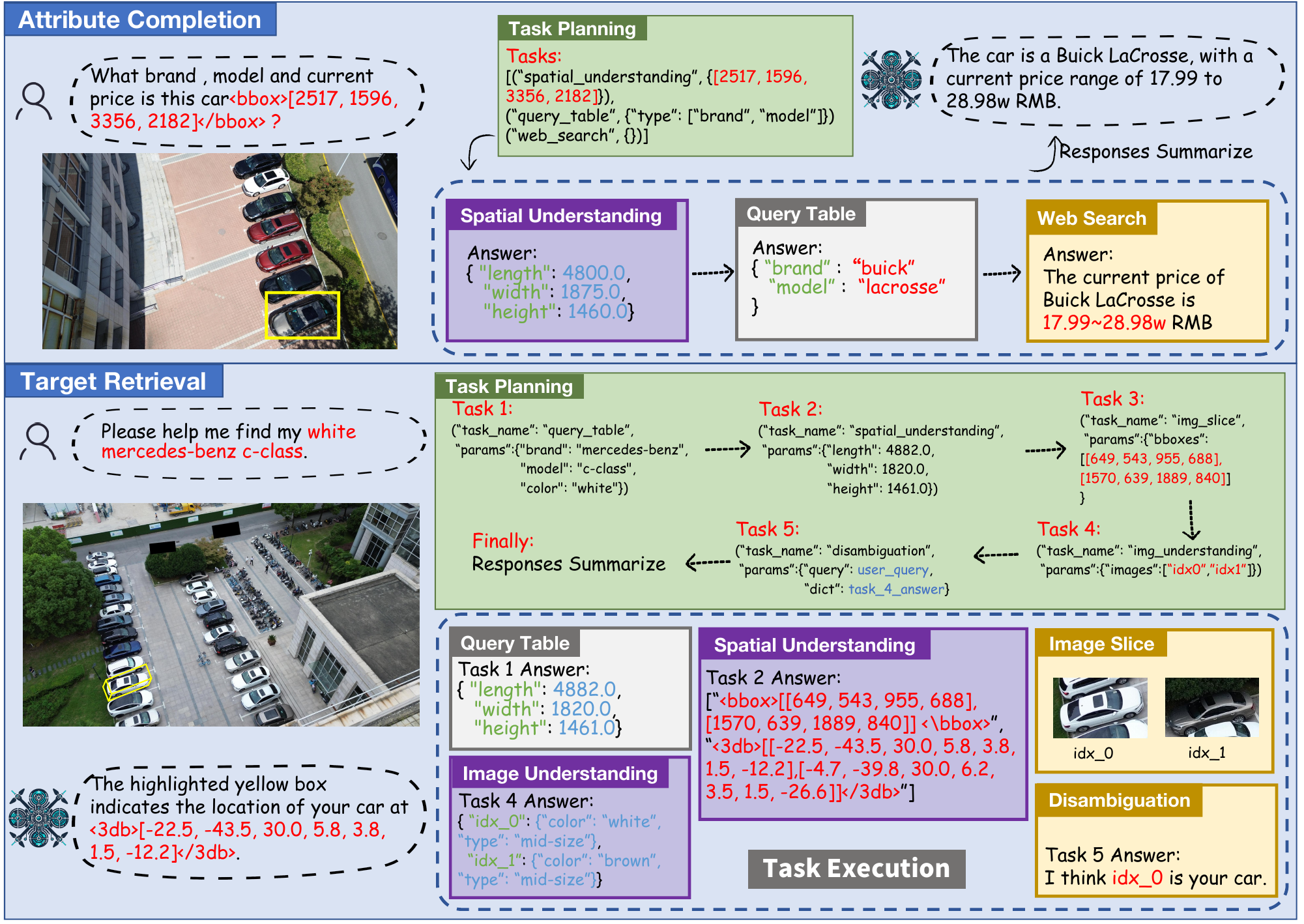}
    \end{center}
    \vspace{-0.5cm}
    \caption{Workflows for Vehicle Attribute Recognition, Zero-Shot Attribute Recognition and Target Retrieval Tasks.}
    \label{fig:agent}         
\end{figure*}

\section{Spatially-Aware Aerial Agent}

To highlight the practical utility of spatially-aware VLMs in remote sensing, we designed an aerial agent for vehicle recognition and retrieval. To achieve these two functionalities, we define the tools listed in the Tab. \ref{tab:tools}. Among them, image understanding and spatial understanding require invoking the VLM, while the web search tool primarily retrieves dynamic information, such as vehicle prices.

\begin{figure}[!t]
	\begin{center}            
        \includegraphics[width=1\linewidth]{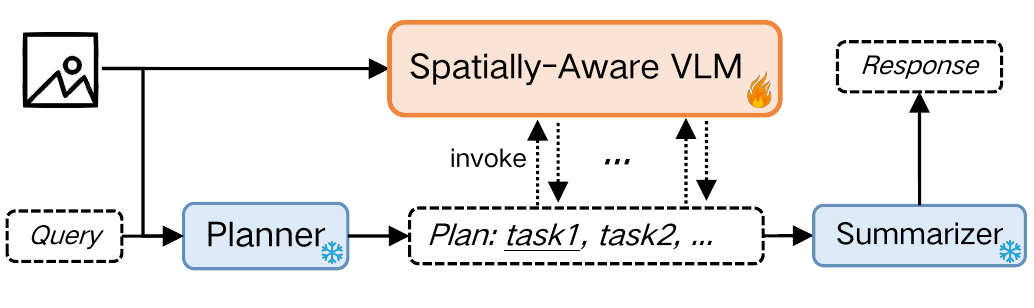}
    \end{center}
    \vspace{-0.5cm}
    \caption{Framework of AirSpatialBot.}
    \vspace{-0.5cm}
    \label{fig:agent_framework}         
\end{figure}

\subsection{Framework}

As shown in Fig. \ref{fig:agent_framework}, the AirSpatialBot is driven by our spatial-aware VLM, seamlessly integrated with an LLM that functions as both the Planner and Summarizer. The LLM requires no additional training and can be guided flexibly through in-context learning by providing a few illustrative examples. When a user asks a question, both the image and the query are first sent to the LLM. The LLM formulates a plan containing multiple tasks based on the query. The agent subsequently executes each task step-by-step according to the generated plan. If a task involves image understanding or spatial understanding, the agent automatically invokes our VLM. The VLM processes the image and the query provided by the agent to generate the corresponding output, which is then returned to the agent. The agent uses the returned value to continue executing the subsequent tasks. Once all tasks are completed, the LLM is called again to summarize and provide the final answer. This is the operational principle of the agent framework. In the following subsection, we will delve into the specific workflows of each representative query.

\subsection{Workflow}

The workflow of our assistant consists of three main steps: plan generation, execution and responses summarize. First, the LLM takes the language-based query as input, flexibly generating the corresponding plan based on different questions. Next, the framework sequentially invokes the VLM and the tools available to execute the plan. Finally, the LLM synthesizes the intermediate results obtained throughout the execution phase to construct a concise, informative final response. This entire process is fully automated, eliminating the need for human intervention. Fig. \ref{fig:agent} illustrates three representative workflows of our aerial agent: attribute recognition, zero-shot attribute recognition, and target retrieval. Below, we will introduce each of them in detail.

\paragraph{Attribute Recognition} It involves identifying specific attributes of a vehicle based on the 2D location provided by the user within an aerial image. For attributes with limited categories and abundant training samples, such as color or type, this can be achieved by fine-tuning the VLM. This workflow generates tasks that only involve image understanding, making it the simplest workflow.

\paragraph{Zero-Shot Attribute Recognition} For attributes such as brand and model, which have numerous categories and insufficient training samples, fine-tuning the VLM becomes challenging. To address this, we leverage the spatial perception capabilities of the VLM as an alternative approach. Specifically, the LLM guides the process by first identifying the vehicle's length, width, and height. It then matches these dimensions to the closest brand and model in a vehicle parameter table. For dynamic attributes like price, which fluctuate with the market, we can retrieve the information by searching the web, such as calling the API of the DCD. This workflow is equally applicable to vehicles that the model has never encountered before, which is why we refer to it as zero-shot attribute recognition. This is particularly advantageous in the highly commercialized automotive industry, where new brands and models are introduced at a rapid pace. Using the proposed method to identify new vehicles only requires a simple update to the parameter table, avoiding the need for supervised learning methods that involve collecting, annotating data, and retraining the model from scratch.

\paragraph{Target Retrieval}

In urban security scenarios, police may need to quickly locate a specific vehicle within a large-scale outdoor scene based on attribute information provided by witnesses. This is where the target retrieval workflow proves useful. Unlike zero-shot attribute recognition, this task first retrieves the target vehicle's length, width, and height information from the table and then uses these dimensions to locate the vehicle from 3D scenes. Thanks to the two-stage training strategy adopted by AirSpatialBot, we can output not only the 2D position of the target but also its 3D position, meeting various user requirements.

\section{Experiment and Analysis}

We adopt the AdamW optimizer with an initial learning rate of 2e-4, weight decay of 0. Gradient clipping with a maximum norm of 1.0 is applied. The learning rate scheduler employs a linear decay strategy with a warm-up ratio of
0.03. Following the experimental configuration of GeoChat~\cite{kuckreja2024geochat}, we set the LoRA rank to 64, with an alpha value of 16 and a dropout rate of 0.05. LoRA modules are specifically applied to the linear layers of connectors and LLMs. We utilize the numerical precision of FP16. The global batch size is set to 128, and the training process runs for a total of 
5 epochs. All experiments were performed on
8 NVIDIA V100 GPUs.


\begin{table*}[!t]
\caption{Performance (Acc@0.5\%) comparison on six existing HBB-based remote sensing visual grounding benchmarks. $\dagger$ denotes that the experimental results are cited from the original paper.} 
\setlength{\tabcolsep}{8pt}
\resizebox{\textwidth}{!}{
\begin{tabular}{lcccccccccc}
\toprule

\multirow{2}{*}{Model} & \multicolumn{2}{c}{DIOR-RSVG} &  \multicolumn{2}{c}{RSVG} &  \multicolumn{2}{c}{OPT-RSVG} & GeoChat  & VRSBench  & RSVG-HD & \multirow{2}{*}{AVG}\\
\cline{2-10}
 & Test & Val & Test &  Val & Test &  Val & Test & Test & Test\\
\midrule   
\multicolumn{11}{l}{\hfill \textit{Specialized Models} } \\
\midrule
MGVLF $^\dagger$\cite{zhan2023rsvg} & \textcolor{gray}{76.78}   & -    &  - & -    & \textcolor{gray}{72.19}   & -  & -  & -  & \textcolor{gray}{50.70}  & -    \\
GeoVG $^\dagger$ \cite{sun2022visual} & -   &  -   & \textcolor{gray}{59.40}  & \textcolor{gray}{58.20}    & -   & -  & -  & -  & -  & -   \\

LQVG $^\dagger$ \cite{lan2024lqvg} &  \textcolor{gray}{83.41}  &  -   & -  & -   & -   & -  & -  & -  & \textcolor{gray}{87.37}  & -   \\

LPVA$^\dagger$\cite{li2024lgpa} &  \textcolor{gray}{82.27}  &  -   & -  & -   & \textcolor{gray}{78.03}  & -  & -  & -  & -  & -   \\

\midrule
\multicolumn{11}{l}{\hfill \textit{Large Vision-Language Models} } \\
\midrule
InternVL2-8B \cite{chen2023internvl} & 14.42  & 12.99  & 0.16 & 0.67  & 12.51  & 11.42 & 9.91   & 5.47  & 1.00 & 7.62    \\
InternVL2-40B \cite{chen2023internvl} & 15.06  & 14.87  & 0.41  & 0.67 & 9.29  & 8.87 & 21.13  & 13.64  & 0.80 & 9.42       \\
Qwen-VL \cite{bai2023qwen} & 32.22  & 32.01   & 2.04  & 4.66  & 30.16  & 29.78 & 35.36   & 31.07  & 10.42  & 23.08    \\
Qwen2-VL \cite{wang2024qwen2} & 44.25  & 43.32 & 20.13  & 19.15  & 37.24 & 36.77 & 30.92   & 32.88  & 31.86 & 32.95     \\
GeoChat \cite{kuckreja2024geochat} & 24.05   & 23.35   & 2.04  & 3.08 & 16.07   & 15.27   & 22.74   & 11.52 & 6.61 & 13.86   \\
LHRS-Bot \cite{muhtar2024lhrs} & 17.59   & 17.04  & 1.56  & 0.95  & 3.69  & 3.20 & 3.25  & 1.19  & 1.08 & 5.51     \\
\rowcolor{myblue}Ours & \textbf{77.41}  & \textbf{77.16} & \textbf{24.93}  & \textbf{24.98} & \textbf{77.31}  & \textbf{77.26}  & \textbf{67.59} & \textbf{63.31} & \textbf{40.08} & \textbf{58.89}   \\
\bottomrule
\end{tabular}}
\label{tab:rec_overall}
\vspace{-0.3cm}
\end{table*}

\subsection{ Spatial Grounding}

\paragraph{Settings}
This task involves locating the corresponding vehicle in the image using spatially-informed referring descriptions. Due to the inability of most existing VLMs to directly produce 3DBB, we utilize HBB for performance comparison. We adhere to standard evaluation protocols~\cite{pang2024h2rsvlm,li2024vrsbench} and assess the spatial grounding task using the Acc@0.5 metric. For GeoChat~\cite{kuckreja2024geochat}, we convert its output OBBs into corresponding HBBs for consistency.

\paragraph{Results on AirSpatial-G} 
Tab. \ref{tab:ae_2} presents the scores of VLMs on four spatial grounding subtasks: the absolute and relative sizes of vehicles, and the absolute and relative distances of vehicles from the drone. AirSpatialBot consistently achieves the highest performance across all spatial grounding scenarios, yet the achieved scores highlight the inherent challenges of these tasks, indicating significant room for improvement. Additionally, we observed that absolute spatial descriptions generally pose greater difficulty than relative descriptions for all evaluated models. Interestingly, GeoChat, a VLM specifically tailored for remote sensing applications, exhibits inferior results compared to general-purpose VLMs. This outcome suggests that the currently available volume and diversity of remote sensing-specific training data remain insufficient to endow specialized VLMs with robust generalization and spatial reasoning capabilities.

\begin{table}[!t]
\caption{Performance (Acc@0.5\%)  comparison of  spatial grounding task on our AirSpatial-G.}
  \centering
  \setlength{\tabcolsep}{2pt}
  \resizebox{\linewidth}{!}{
  \begin{tabular}{l|cccc|c}
    \toprule
      Method & Abs. Size & Rel. Size & Abs. Distance & Rel. Distance & AVG \\ \hline
    InternVL2  &  0.58 & 3.46 & 0.74 & 8.10 & 3.22 \\
    Qwen-VL  & 0.92 & 6.52 & 2.20 & 10.64 & 5.07 \\
    Qwen2-VL  & 3.60 & 8.47 & 3.72 & 18.58 & 8.59 \\
    GeoChat  & 0.41 & 0.65 & 0.36 & 2.08 & 0.88 \\
    \rowcolor{myblue}AirSpatialBot  & \textbf{6.23} & \textbf{12.54} & \textbf{6.43} & \textbf{26.65} & \textbf{12.96} \\
    \bottomrule
  \end{tabular}}
    
    \label{tab:ae_2}
\vspace{-0.3cm}
\end{table}

\begin{figure*}[!t]
\vspace{-0.2cm}
	\begin{center}             
        \includegraphics[width=1\linewidth]{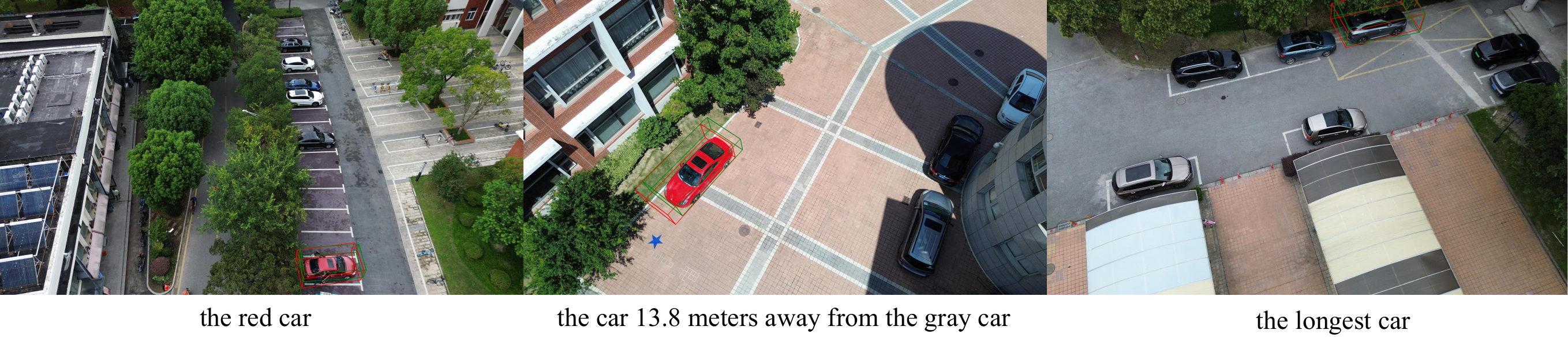}
    \end{center}
\vspace{-0.5cm}
    \caption{Visualizations of AirSpatialBot on AirSpatial-G with 3DBB. The green boxes indicate ground truth, while the red boxes represent predictions.}
    \label{fig:viz_3dbb}        
\end{figure*}


\paragraph{Results on other RSVG benchmarks} 

To demonstrate that our model also exhibits strong performance in 2D remote sensing visual grounding tasks, we conduct evaluations on six existing RSVG benchmarks. Table \ref{tab:rec_overall} presents the performance metrics of 13 models, including specific models, general VLMs, remote sensing VLMs, and the VLM obtained from the first stage of our two-stage training.  The experimental results demonstrate that our model significantly outperforms existing VLMs in remote sensing visual grounding tasks, laying a solid foundation for the subsequent second-stage 3D knowledge transfer. It is worth noting that our VLM still lags behind proprietary models on some benchmarks, which is largely attributed to the resolution of the input images. Since images are resized to the model default resolution during input (336$\times$336), small objects in remote sensing data become difficult to locate accurately.

\paragraph{Visualization Examples}

Fig. \ref{fig:viz_3dbb}  illustrates three examples demonstrating the 3D grounding capability of our spatially-aware VLM. The model is capable of generating 3D bounding boxes to accurately localize target vehicles within a 3D scene. Specifically, the leftmost image primarily evaluates the model’s visual understanding ability, while the rightmost image focuses on its spatial understanding capability. The middle image, on the other hand, simultaneously assesses both visual and spatial understanding, highlighting the model’s capability to integrate these two aspects effectively. It is worth noting that our model is the first remote sensing VLM to simultaneously support both 2D  and 3D grounding.



\begin{figure*}[!t]
\vspace{-0.2cm}
	\begin{center}             
        \includegraphics[width=1\linewidth]{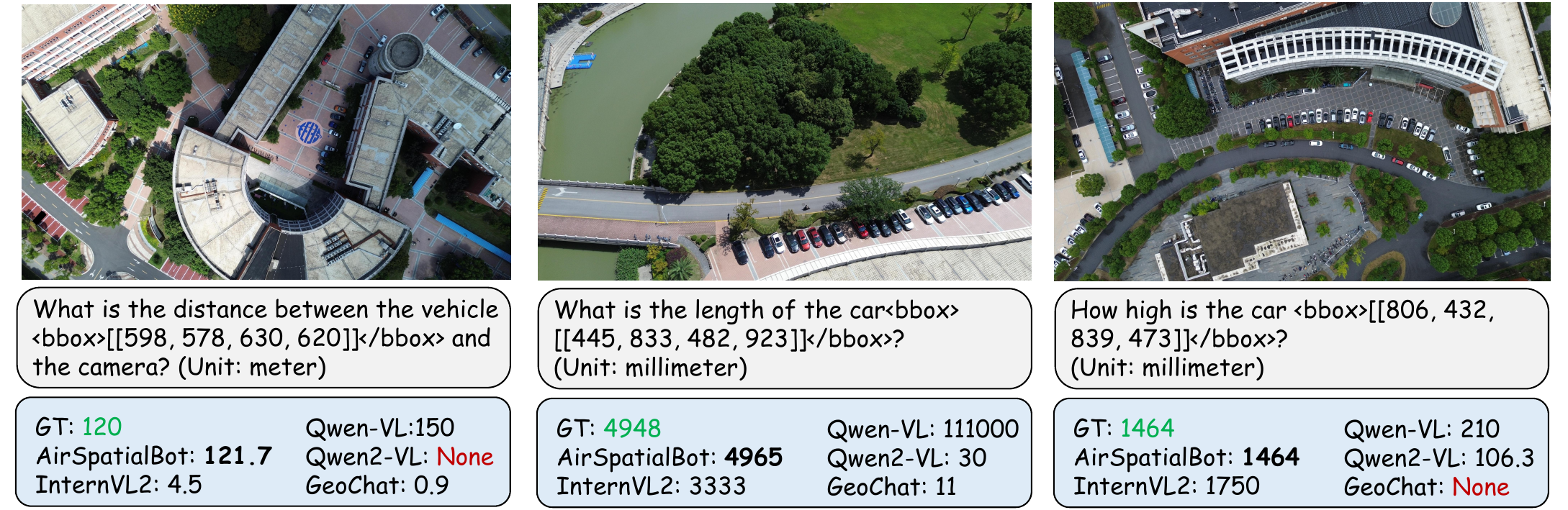}
    \end{center}
\vspace{-0.5cm}
    \caption{Comparison of the SQA Task on AirSpatial-QA. Only the numerical values from the answers have been extracted.}
    \label{fig:viz_sqa}        
\end{figure*}

\subsection{Spatial Question Answering}

\paragraph{Settings}
Following REO-VLM~\cite{xue2024reo}, we employ RMSE, MAE and R-squared as evaluation metrics for regression-based VQA tasks. Since the queries in SQA questions contain target coordinate information, only VLMs with regional captioning capabilities can be evaluated. Therefore, we select four mainstream VLM models with this capability for comparison. Among them, Qwen-VL, Qwen2-VL and InternVL2 are general-purpose VLMs, while GeoChat is a remote sensing VLM.

\begin{table}[!t]
  \caption{Performance comparison of SQA task on AirSpatial-QA.}
  \setlength{\tabcolsep}{10pt}
  \resizebox{\columnwidth}{!}{
  \begin{tabular}{lccc}
    \toprule
     Model  & MAE $\downarrow$ &  RMSE$\downarrow$ & R-squared $\uparrow$ \\
    \midrule
     InternVL2  & 1343.07 & 2083.38 & -0.57\\
     Qwen-VL  & 1467.40 & 2223.76 & -0.64  \\
     Qwen2-VL  & 2028.80 & 2585.33 & -1.38  \\
     GeoChat  & 1551.04 & 2296.48 & -0.82  \\
\rowcolor{myblue}AirSpatialBot & \textbf{103.80} & \textbf{216.19} & \textbf{0.99} \\
    \bottomrule
  \end{tabular}}
    
    \label{tab:ae_4-1}
\end{table}

\begin{table}[!t]
  \caption{Fine-Grained performance (MAE) comparison of SQA task.}
  \setlength{\tabcolsep}{4pt}
  \resizebox{\columnwidth}{!}{
  \begin{tabular}{lccccc}
    \toprule
     Model  & Depth & Distance & Length & Width & Height  \\
    \midrule
     InternVL2  & 101.90 & 103.98 & 3725.97 & 1466.77 & 1455.05 \\
     Qwen-VL  & 229.27 & 165.15 & 3729.18 & 1237.57 & 1067.04\\
     Qwen2-VL  & 87.14 & 232.70 & 4418.97 & 1728.30 & 1430.40 \\
     GeoChat  & 76.64 & 60.24 & 4597.29 & 1634.39 & 1404.93 \\
    \rowcolor{myblue}AirSpatialBot & \textbf{2.90} & \textbf{3.08} & \textbf{210.66} & \textbf{45.67} & \textbf{99.67}  \\
    \bottomrule
  \end{tabular}}
    
    \label{tab:ae_4}
\end{table}

\paragraph{Results}

As shown in Tab. \ref{tab:ae_4-1}, AirSpatialBot significantly outperforms the other four VLMs across all three evaluation metrics. AirSpatialBot achieves a 0.99 R-squared score on the SQA task, significantly outperforming existing VLMs such as InternVL2 (R-squared = -0.57) and GeoChat (R-squared = -0.82). InternVL2 ranks second but still shows a significant gap compared to AirSpatialBot. Although GeoChat is a specially trained remote sensing VLM, it does not demonstrate a clear advantage over general-purpose VLMs. Fig. \ref{fig:viz_sqa} provides a more intuitive visualization of the performance gap among different models in the SQA task. It is important to note that SQA specifies the target using textual bounding boxes without modifying the image.

To further investigate the spatial perception capabilities of VLMs, we present in Tab. \ref{tab:ae_4} the MAE score of each model in five fine-grained tasks: estimating the depth, distance, length, width, and height of specific vehicles. AirSpatialBot achieves the best performance across all five subtasks. Althoughugh GeoChat ranks secobehind AirSpatialBot in depth and distanceistance estimation, it performs poorly in estimating the 3D dimensions of vehicles. Our findings also indicate that, for all VLMs, estimating vehicle dimensions is considerably more challenging than estimating depth and distance. Among the dimension estimation tasks, predicting vehicle length proves to be the most difficult. Enhancing the VLM's ability to perceive vehicle dimensions could further improve agent's overall capabilities, making it even more effective in vehicle attribute recognization and
retrieval.

\begin{table*}[!t]
\caption{Performance comparison of attribute recognization and retrieval on our AirSpatialBench.}
  \centering
  \setlength{\tabcolsep}{12pt}
  \resizebox{\textwidth}{!}{
  \begin{tabular}{l|cccccc|c}
    \toprule

    \multirow{2}{*}{Method} & \multicolumn{6}{c|}{Attribute Recognization} &  \multirow{2}{*}{Target Retrieval} \\
    \cline{2-7}
      & Brand & Model & Powertrain & Price & Doors &  AVG &   \\  \hline
    InternVL2   & 5.97 & 0.75 & 0.00 & 7.84 &  25.00 & 7.91 & 0.00 \\
    Qwen-VL & 3.36 & 0.75 & 0.00 & 0.00  & 25.00 & 5.82 & 0.00\\
    Qwen2-VL   & 0.00 & 0.00 & 0.00 & 0.00 & 0.00  & 0.00 & 0.00\\
    GeoChat  & 0.00 & 0.00 & 0.00 & 0.00  & 25.00  & 5.00 & 0.00 \\
\rowcolor{myblue}AirSpatialBot & \textbf{7.84} & \textbf{1.49} & \textbf{78.95} & \textbf{12.69} & \textbf{41.67} & \textbf{28.53} & \textbf{29.74}\\
    \bottomrule
  \end{tabular}}
    
    \label{tab:ae_1}
\end{table*}

\subsection{Vehicle Attribute
Recognization and Retrieval}

\paragraph{Settings}

Similar to previous studies~\cite{kuckreja2024geochat,li2024vrsbench}, we use accuracy as the evaluation metric for the attribute recognition task. For the attribute recognition task, which is essentially a VQA task, we use accuracy as the evaluation metric. For the target retrieval task, we locate vehicles in 3D scenes using 3DBB. Therefore, we adopt the commonly used BEV IoU (Bird’s Eye View Intersection over Union) in the autonomous driving domain~\cite{wang2024omnidrive} to calculate accuracy. A prediction is considered correct if the overlap between the two 3DBBs exceeds 0.25; otherwise, it is considered incorrect.

\paragraph{Results}

Our study pioneers the innovative application of VLMs' spatial reasoning capabilities to achieve zero-shot, fine-grained aerial vehicle attribute recognition. Unlike previous works on vehicle attribute recognition based on remote sensing images, which primarily focused on basic attributes such as color and type, our approach extends the scope to fine-grained attributes, including brand, model, powertrain, price, seating capacity, and the number of doors. Furthermore, we demonstrate that vehicles in 3D scenes can be localized based on their brand and model. Tab. \ref{tab:ae_1} compares the performance of our method with existing VLMs in these two tasks. Our model achieved the highest average score on the attribute recognition task and is the only VLM that supports the target retrieval task that requires 3DBB output. Notably, we observed that Qwen2-VL exhibited a decline in performance compared to Qwen-VL in the attribute recognition task, rather than showing an expected improvement. This indicates that even though a VLM may demonstrate significant advancements on general benchmarks, its performance on remote sensing-specific benchmarks does not necessarily improve. This reveals a shortcoming in the generalization capability of current VLMs when applied to domain-specific tasks. The experimental results indicate that existing models lack the capability to recognize fine-grained vehicle attributes from aerial imagery.

\begin{table}[!t]
\caption{Ablation Study of AirSpatialBot on spatial grounding task.}
  \centering
  \setlength{\tabcolsep}{16pt}
  \resizebox{\columnwidth}{!}{
  \begin{tabular}{lcccc}
    \toprule

    \multirow{2}{*}{Options} & 
    \multicolumn{2}{c}{BEV}  \\
    \cline{2-3}
    & Acc@0.25 & Acc@0.5  \\
    \midrule
    LLaVA-1.5-7B & 10.67 & 3.86   \\
    \midrule
    + 2D Pretrain & 15.68 & 6.37   \\
    + Multiple Signals & 25.43 & 13.33   \\
    + GML & 25.54 & 13.24  \\
    \rowcolor{myblue} + ASL & \textbf{28.88} & \textbf{15.51}  \\
    \bottomrule
  \end{tabular}}
    
    \label{tab:me_3_1}
\end{table}

\subsection{Ablation Studies}

Tab. \ref{tab:me_3_1} demonstrates the complete process of improvements made to the spatially-aware VLM used in our AirSpatialBot. 2D Pretrain represents the Image Understanding Pre-training in the two-stage training process. The improvements it achieves confirm that 2D knowledge can indeed benefit 3D tasks. Multiple Signals refers to providing three types of supervision signals—HBB, OBB, and 3DBB—simultaneously for each target during the second stage. This allows the model to develop a more diverse and comprehensive understanding of the target's location. ASL and GML explicitly enable the model to learn the relationship between 2D and 3D signals, allowing it to more efficiently absorb and leverage existing 2D knowledge and transfer it to 3D tasks.

\begin{table}[!t]
\caption{Ablation Study of LLMs in AirSpatialBot.}
  \centering
  \setlength{\tabcolsep}{6pt}
  \resizebox{\columnwidth}{!}{
  \begin{tabular}{lcc}
    \toprule

    Options & 
    Attribute Recognization & Target Retrieval\\
    \midrule
    No extra LLM & 0.00  &  0.00 \\
    + GPT-3.5-Turbo & 24.76   &  26.23 \\
     + DeepSeek-v3 & 28.32  & 28.69 \\
    \rowcolor{myblue} + GPT-4o & \textbf{28.53}  & \textbf{29.74}  \\
    \bottomrule
  \end{tabular}}
    
    \label{tab:me_3_3}
\end{table}

Tab. \ref{tab:me_3_3} compares the impact of different LLMs on the performance of our aerial agent. "No extra LLM" means that no additional LLM is used; instead, our spatial-aware VLM is directly employed to generate the plan. However, we found that this approach performs poorly, possibly because the visual processing capability affects the inherent logical reasoning ability of the LLM. Therefore, we implemented a dual-model approach to collaboratively operate within this framework: the LLM is responsible for logical reasoning, while the VLM focuses on interpreting images and spatial information. Next, we evaluate three state-of-the-art LLMs. GPT-4o achieves the best results, while DeepSeek~\cite{liu2024deepseek} follows closely with a small performance gap. This is because the task of planning itself is not particularly difficult and primarily requires a basic level of in-context learning ability. However, the overall scores remain relatively low, which can be attributed to the limitations of the spatial perception model. Therefore, to further enhance the performance of AirSpatialBot, the key lies in improving the spatial perception capabilities of the VLM.

\section{discussion}

Although our study primarily focused on vehicles, AirSpatialBot's proposed framework can be naturally extended to other types of ground targets, such as aircraft and ships. Future research will investigate its applicability and performance in more dynamic and complex scenarios, including disaster response and urban surveillance. In practical applications, AirSpatialBot is particularly well-suited for scenarios requiring high mobility and flexibility. For relatively static environments, such as parking lots, fixed cameras typically suffice to assist users in locating their vehicles, reducing the necessity for drone-based approaches. However, during critical emergency situations (e.g., disaster response, search-and-rescue operations), conventional fixed-camera systems and network infrastructure often become compromised or inoperable due to structural damage, environmental conditions, or power outages. Under these challenging circumstances, AirSpatialBot demonstrates substantial advantages by utilizing drones to rapidly identify and localize critical targets, as well as retrieve essential real-time information. Such capabilities significantly enhance situational awareness, enabling swift and informed decision-making processes. Consequently, AirSpatialBot is particularly valuable for missions where rapid deployment, adaptability, and resilience to infrastructure disruptions are crucial. By integrating autonomous aerial perception and intelligent target recognition, AirSpatialBot enhances operational effectiveness in complex and dynamic environments.

\section{conclusion}

In this paper, we introduced AirSpatial, a novel aerial dataset designed to enhance spatial perception in remote sensing vision-language models (VLMs). Using a two-stage training strategy, we transferred image comprehension abilities into robust spatial reasoning capabilities. Additionally, we proposed AirSpatialBot, a spatially-aware aerial agent for fine-grained vehicle attribute recognition and retrieval. Extensive experiments validated the effectiveness of our approach, highlighting both challenges and opportunities for advancing spatially-aware VLMs in remote sensing.



%




\ifCLASSOPTIONcaptionsoff
  \newpage
\fi



\bibliographystyle{IEEEtran}
\bibliography{Mendeley_Tarot}
\end{document}